# VISUAL ODOMETRY for MOVING RGB-D CAMERAS

Afonso Fontes
Universidade de Fortaleza
(UNIFOR)
afonsohfontes@gmail.com

José Everardo Bessa Maia
MACC
Universidade Estadual do Ceará
(UECE)
jebmaia@gmail.com

*Abstract*—Visual odometry is the process of estimating the position and orientation of a camera analyzing the images associated to it. This paper develops a quick and accurate approach to visual odometry of a moving RGB-D camera navigating on a static environment. The proposed algorithm uses SURF (Speeded Up Robust Features) as feature extractor, RANSAC (Random Sample Consensus) to filter the results and Minimum Mean Square to estimate the rigid transformation of six parameters between successive video frames. Data from a Kinect camera were used in the tests. The results show that this approach is feasible and promising, surpassing in performance the algorithms ICP (Interactive Closest Point) and SfM (Structure from Motion) in tests using a publicly available dataset.

## I. INTRODUCTION

Visual Odometry is the ability to track the motion of an agent over time using only visual information. Its global position and orientation can be obtained from the pose of the associated camera. Also known as VO, Visual Odometry is widely used for performing precision tasks in many important applications in the fields of robotics and machine vision, especially when no other sensing system is available, as [1] and [2] demonstrate.

The contribution of this paper is to provide a fast and precise odometry using information obtained through a single RGB-D camera. With low cost and high potential, these type of sensors were designed for entertainment purposes. At video frame rates and VGA resolution, a RGB-D camera captures synchronized color images and depth maps. This allows combining techniques in order to more accurately estimate the pose difference between two readings. Figure 1 shows two frames and a dense 3D representation of the environment, which can be created after computing the new camera pose using the algorithm proposed on this paper.

This paper is organized in four main sections. A brief review of other VO approaches and related work will be discussed on Section II. Section III presents the sequence of methods used to estimate movement and their functionality. Section IV compares the performance of the proposed method with SfM and ICP, other two VO approaches frequently used on many applications. This evaluation is made on sets of sequenced RGB-D captures publicly released by Sturm [3], whom also provides the real trajectory of the moving sensor acquired from an external high accuracy motion capture system. Therefore, this work is expected to encourage and cooperate with future systems based on RGB-D readings yet to be developed.

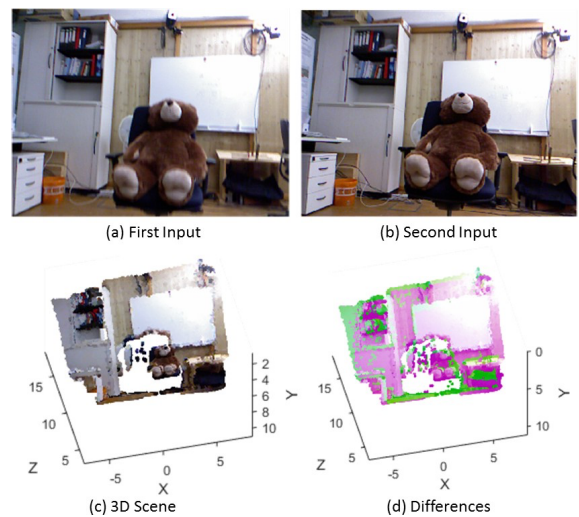

Fig. 1. The proposed approach estimates the camera pose using two frames, (a) and (b), to estimate the rigid body transformation between them. With the color and depth informations provided by the sensor, it is possible to generate a 3D scene (c), which is the sum of the green (a) and magenta (b) depth readings seen in (d).

## II. RELATED WORK

Methods of Visual Odometry based only on colored (RGB) images, such as Structure from Motion (SFM) performed by Benseddik [?], typically identify points with strong features in a reference image and esteem their locations in a subsequent image to then, given a camera model, estimate their depth by re-projection. The motion made by the camera can be computed by minimizing the differences between these two sparse sets of points. On the other hand, there are also approaches based only on depth readings, such as Interactive Closest Points (ICP), fisrt introduced by [4] and recently used on the contributions of [5] and [6]. In this case, the motion is computed using all the points available on both captures.

There are many methods available to identify strong features on RGB images in order to locate relevant points, among them,

the commonly used SURF, abbreviation of Speeded Up Robust Features [?], as used in the works of [7] [8]. SURF is called a blob detector and its main idea is to identify sets of pixels of the same tonality. Once identified on the first image, the points must now be tracked into the subsequent frame using Feature Matching [9], which returns the possible new coordinate of each given point. These correspondences might not be accurate due to the similarity of spots over the environment, resulting on some fake correlations. In order to remove these wrong data and increase the overall robustness of the process, it is required to filter out those points where the variation of location in the images did not follow a standard when compared to all correlated points. For this problem, the most used method is Random Sample Consensus (RANSAC) [10], which filters inconsistent parameters in many applications.

For each captured frame, a RGB-D camera outputs a color image (RGB) and a depth map (D). This enables us to work with all the techniques previously mentioned separately and/or merge them, creating a new approach, as did Milella [11], which the contribution consists of finding and tracking points over two RGB images, reading their depths directly from the Depth image and finally using ICP to compute the motion.

The method proposed in this paper is mainly based on the contributions of [12] and [13]. It consists of four main stages of processing, as illustrated in Figure 2. First SURF is executed in a couple of RGB images to extract relevant points of each frame. On the next two blocks these points are matched and then filtered using Feature Matching and RANSAC respectively. Finally, in order to estimate the movement between two frames, the last block reads the depth of these points directly from the sensor data and minimize the difference between them using Single Value Decomposition (SVD), which the step by step is documented in the next section of this paper. It is also possible to estimate the camera global trajectory by concatenating all transformations computed.

As a result of our algorithm provides a sequence of poses associated with each pair of frames as do the contributions of [14] and [15]. The estimated movement can be used in various applications such as reconstruction of 3D scenes ( [16], simultaneous localization systems and mapping (SLAM) [17], or even for medical purposes, such as documented by Spyrou [18], which uses Visual Odometry on images captured on a endoscopy exam in order to help on the diagnosis.

## III. PROPOSED ALGORITHM

*1) Overview:* This section presents the approach used to estimate the pose a moving RGB-D camera. Our goal is to compute a rigid body motion, composed of a translation (t) and an rotation (R), that minimizes the difference between two sets of points $P$ and $Q$, as shown in 1. For better understanding, this section is divided into two other subsections. The next subsection reports the process that locate, match and filter relevant points on the RGB images. After that, the process to estimate the motion is shown.

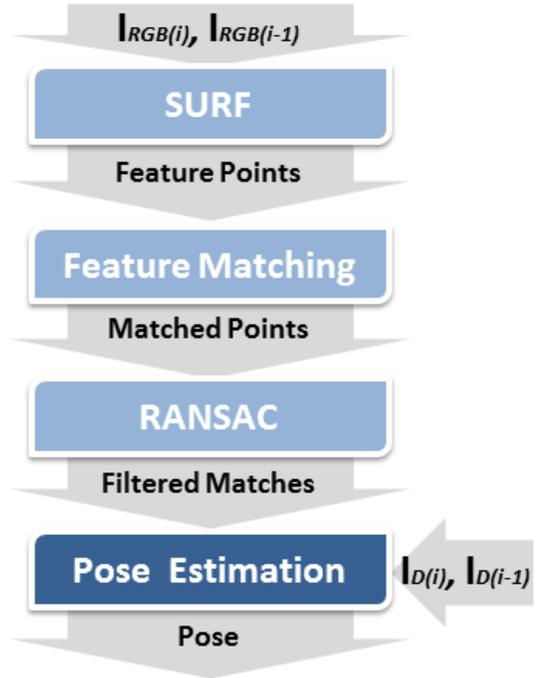

Fig. 2. Flow chart of techniques applied to estimate the camera motion from a pair of RGB images, $I_{RGB}(i)$ and $I_{RGB}(i-1)$, and their respective depth maps, $I_D(i)$ e $I_D(i-1)$.

$$(R, t) = min \sum_{i=1}^{n} |(R \times p_i + t) - q_i|^2 \quad (1)$$

*2) Finding and Matching Visual Features:* SURF is a robust technique for detecting key points in images, especially when they are subjected to change as rotation, shift and scale. First presented by Bay in 2006 [19], SURF is often used in computer vision tasks as object recognition and 3D reconstruction [20]. Thus the first step of the proposed method is to extract feature points using SURF. In order to measure the displacement of a point it is necessary to locate it on the second frame using Feature Matching [9]. Figure 3 shows the results of these two steps on consecutive frames. The correspondences are linked by a yellow line. Note that there are points which do not follow the standard displacement between the images and therefore, in order to achieve robustness, they must be filtered before calculating the pose.

The RANSAC algorithm proposed by Fischler [10], works on the assumption that the data set in question has inliers and outliers, thus values within a range of acceptance can be filtered from within noisy data. In the context of this paper, RANSAC is used to filter sets of points (coordinates) that do not follow a displacement between the images. The result of this process can be clearly seen on Figure 4. Filtering allows to work only with points which probably have the same rotation and translation thus pose estimation calculation tends to be more accurate. In summary, accuracy increases as we remove false matches.

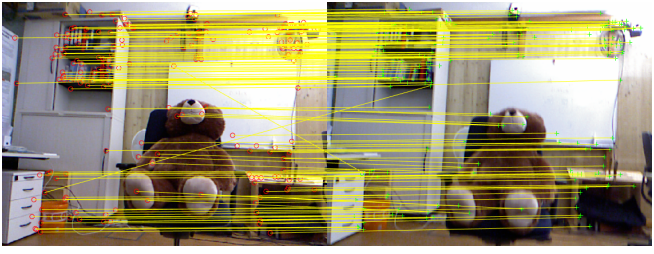

Fig. 3. Feature points located in two consecutive frames using SURF and matched using Feature Matching.

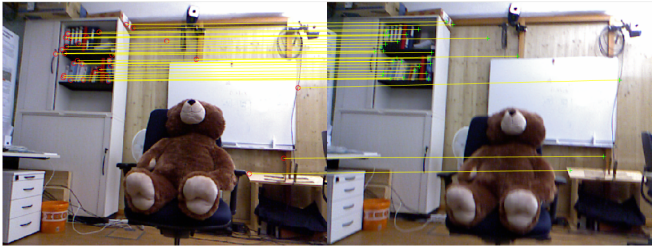

Fig. 4. Matched points filtered using RANSAC.

Once located, matched and filtered, the two sets of points are now ready to proceed to the main step of the whole process: Pose Estimation. The third coordinate of these relevant points can be directly obtained from the depth readings of each frame. Figure 5 illustrates in three dimensions the depths referring to the fisrt input of Figures 3 and 4.

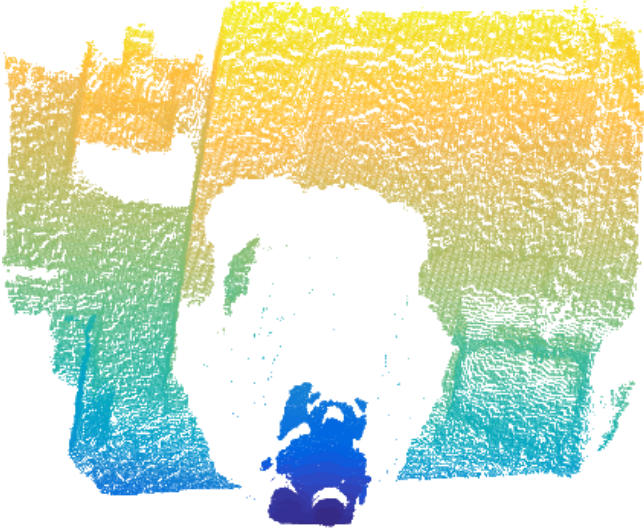

Fig. 5. Three-dimensional representation of a depth reading acquired from a RGB-D camera.

*3) Pose Estimation:* The difference between the current and the last camera poses can be computed trough the calculation of a rigid body transformation between these two frames. Obviously, to estimate the current pose of a moving camera, we add up all the transformations previously calculated. McCarthy [21] explains that it is possible to represent a rigid transformation by a translation vector ($t$) and a rotation matrix ($R$). In this work, t is equal to Cartesian coordinates x, y and z, while the 3x3 identity matrix R represents a zero rotation in all axes. According to McCathy, rotations along each axis is given by the equations 2, 3 and 4.

$$x : \begin{bmatrix} 1 & 0 & 0 \\ 0 & cos\theta_x & -sin\theta_x \\ 0 & sin\theta_x & cos\theta_x \end{bmatrix} \quad (2)$$

$$y : \begin{bmatrix} cos\theta_y & 0 & sin\theta_y \\ 0 & 1 & 0 \\ -sin\theta_y & 0 & cos\theta_y \end{bmatrix} \quad (3)$$

$$z : \begin{bmatrix} cos\theta_z & -sen\theta_z & 0 \\ sen\theta_z & cos\theta_z & 0 \\ 0 & 0 & 1 \end{bmatrix} \quad (4)$$

Let $P = \{p_1, p_2, ..., p_n\}$ and $Q = \{q_1, q_2, ..., q_n\}$ be two sets of $n$ coordinates $x$, $y$ and $z$, Sorkine [12] proposes five steps to find a rotation $R$ and a translation $t$ to solve the equation 1:

- Compute the centroid of the data sets $P$ and $Q$ given by
$\bar{p} = \frac{\sum_{i=1}^{n} p_i}{n}$ e $\bar{q} = \frac{\sum_{i=1}^{n} q_i}{n}$.
- Compute their centralized vectors
$v1_i = p_i - \bar{p}$ e $v2_i = q_i - \bar{q}$, $i = 1, 2, ...n$.
- Find the covariance matrix $S$ of dimensions $d \times d$, where $d$ is the number of Cartesian coordinates used, 3 in our case, given by the multiplication of the matrixes $M_{v1}$ and $M_{v2}$ of size $d \times n$ that have respectively $v1_i$ e $v2_i$ as columns
$$S = M_{v1} \times M_{v2}^T.$$
- Compute the Single Value Decomposition (SVD) of S, given by $S = U\Sigma V^T$, and finally the rotation $R$ using
$$R = V \begin{bmatrix} 1 & & & & \\ & 1 & & & \\ & & ... & & \\ & & & 1 & \\ & & & & det(VU^T) \end{bmatrix} U^T.$$
- At last, the translation $t$ is given by
$$t = \bar{q} - R\bar{p}.$$

## IV. PERFORMANCE ANALYSIS

Publicly available sets of sequenced RGB-D frames were used to qualitatively evaluate the method proposed in this paper. Released by Sturm [3], the datasets were all recorded using one of the first RGB-D cameras released: the Microsoft Kinect. Along with the images, Proposed Method also provides the actual path taken by the camera (groundtruth) acquired by an external high-precision motion capture system. There are a variety of sequences available to download, however, the performance evaluation in this section uses the sequences nammed **freiburg1/floor** and **freiburg1/teddy** because they contain both motions, translation and rotation, in a typical office environment at different speeds and routes. The RGB-D camera was moved manually in both sequences with the difference that in **freiburg1/floor** it does not lose ground reference in most of its frames.

The tests made consist on going through all the frames estimating the change of camera pose between them. In order to compare the performance, the method described in the previous section of this paper is compared to two other distinct techniques: Interative Closest Points (ICP), which takes into account only depth (D) images, and Structure from Motion (SfM), where only color (RGB) images are required to estimate the pose. This way we intend to compare different approaches to Visual Odometry.

All experiments were performed in the Windows 10 operational system, running on a notebook with a 1.7 GHz Intel Core i3 processor and 4GB of RAM. The system is implemented in Matlab version R2016a using the **Computer Vision System Toolbox**, a critical package that assisted achieving the reported performance.

Even tough the camera pose is composed of three location coordinates and respectively three orientation angles, in the results of the experiments the error is shown in meters, calculated taking into account only the camera's location in the environment, as this is directly affected by its orientation.

From the actual path of the RGB-D camera available in the datasets, it is possible to measure and compare the performance of the Visual Odometry systems mentioned before by both, global position at a given frame and the estimatives calculated between each couple of frames.

The first experiment, illustrated in Figures 6 and 7 uses the sequence of RGB-D images **freiburg1/floor**. The camera navigates through an office environment with the ground as a reference in most frames. It is possible to see on Figure 6 that the ICP algorithm, represented by the 'Depth' title, has the most significant peaks, followed by the proposed method in this article, being SfM the method with more consistently small position errors. As both graphs express the error taking into consideration only the camera position, orientation displacements can be seen indirectly in cumulative graph on Figure 7 which illustrates a difference of position between SfM, that apparently had a good result so far, and the given groundtruth of more than 1 meter over 900 frames, which is the worst result among the compared methods.

In the second experiment, done with the sequence **freiburg1/teddy** and illustrated by Figures 8 and 9, the camera performs more brusque orientation movements and the sequence of frames has no constant reference as the floor, a wall or the ceiling. It is possible to note a better performance of the SfM algorithm, which again has a better constancy of small errors computed frame by frame, seen in Figure 8, followed by the proposed method and finally the ICP algorithm, and furthermore, has the slightest accumulated error over 1300 frames, as shown in Figure 9.

The mean errors of all three methods for both experiments can be seen on Tables I and II. Note that the proposed method has the lowest average error and a processing time relatively low for both experiments.

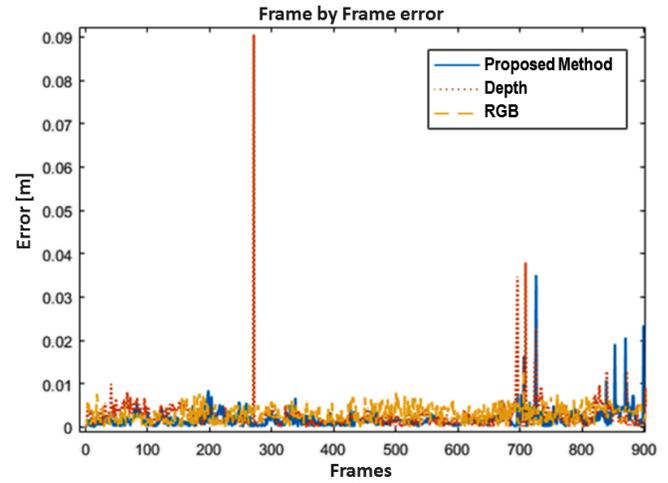

Fig. 6. Comparison of the error using the dataset freiburg1/floor and computed frame by frame. Figure shows the method proposed in this paper against ICP and SfM algorithms, represented respectively by Proposed Method, Depth and RGB titles.

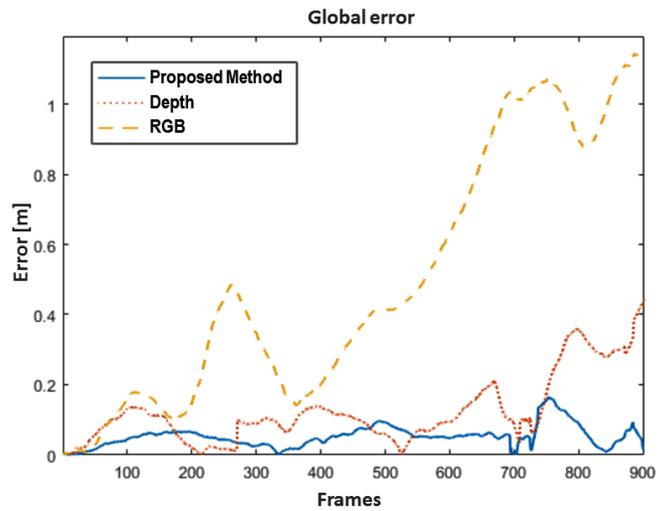

Fig. 7. Comparison of the global position error using the dataset freiburg1/floor. Figure shows the method proposed in this paper against ICP and SfM algorithms, represented respectively by Proposed Method, Depth and RGB titles.

During the execution of the experiments it was possible to notice that ICP algorithm is immune to color variation or light variation in navigated environment, but fails more often when the sensor detects no major variation in depth. While SfM algorithm strongly depends on the color variation from one image to another to improve its accuracy, unable to navigate in dark environments. This difference between the two algorithms can be seen in the passages between the frames 550 and 650 of Figure 9, when the camera navigates in an area with a lot of depth variation the ICP accuracy increases, while over frames

TABLE I
MEAN GLOBAL ERROR AND PROCESSING TIME FOR THE EXPERIMENT FREIBURG1/FLOOR.

| Sequency freiburg1/floor | | |
|---|---|---|
| Algorithm | Mean error | Processing time |
| SfM | 0.51 m | 950 s |
| ICP | 0.12 m | 621 s |
| Proposed Method | 0.05 m | 458 s |

TABLE II
MEAN GLOBAL ERROR AND PROCESSING TIME FOR THE EXPERIMENT FREIBURG1/TEDDY.

| Sequency freiburg1/teddy | | |
|---|---|---|
| Algorithm | Mean error | Processing time |
| SfM | 0.22 m | 1430 s |
| ICP | 0.74 m | 1155 s |
| Proposed Method | 0.13 m | 953 s |

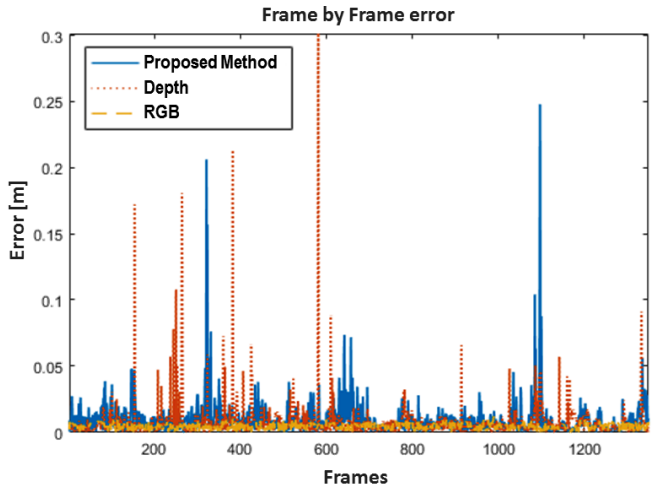

Fig. 8. Comparison of the error using the dataset freiburg1/teddy and computed frame by frame. Figure shows the method proposed in this paper against ICP and SfM algorithms, represented respectively by Proposed Method, Depth and RGB titles.

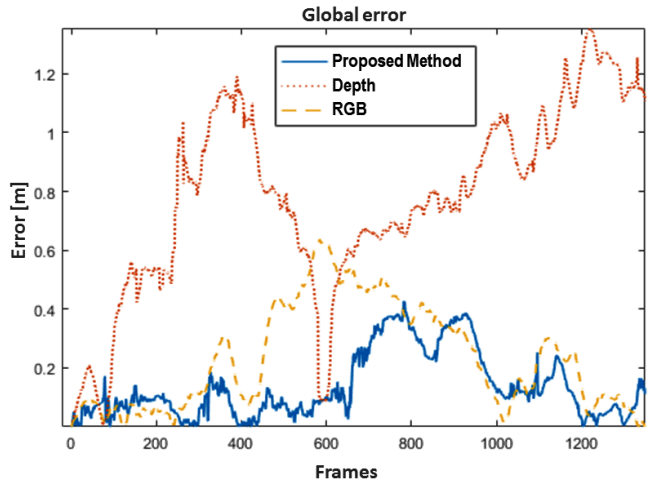

Fig. 9. Comparison of the global position error using the dataset freiburg1/teddy. Figure shows the method proposed in this paper against ICP and SfM algorithms, represented respectively by Proposed Method, Depth and RGB titles.

950 and in 1050 there are a lot of color variation, increasing the accuracy of SfM.

The results of the experiments show that the method proposed in this article has the lowest average error, a relatively low processing time and a satisfactory consistency when compared to the algorithms mentioned.

## V. 3D RECONSTRUCTION OF A SCENE

By concatenating all individual movements, the total path of the camera and of the agent associated to it can be estimated. The association of this trajectory with the data acquired by the RGB-D camera makes possible the creation of dense 3D representations of environments, process illustrated in Figure 1. Figure 10 compares the real trajectory (groundtruth) with the one estimated by the proposed method. The three-dimensional top view representation of the navigated environment is illustrated in in Figure 11 and perspective in Figure 12. Despite the differences, the sequence of computed orientations and translations generated a consistent 3D scene.

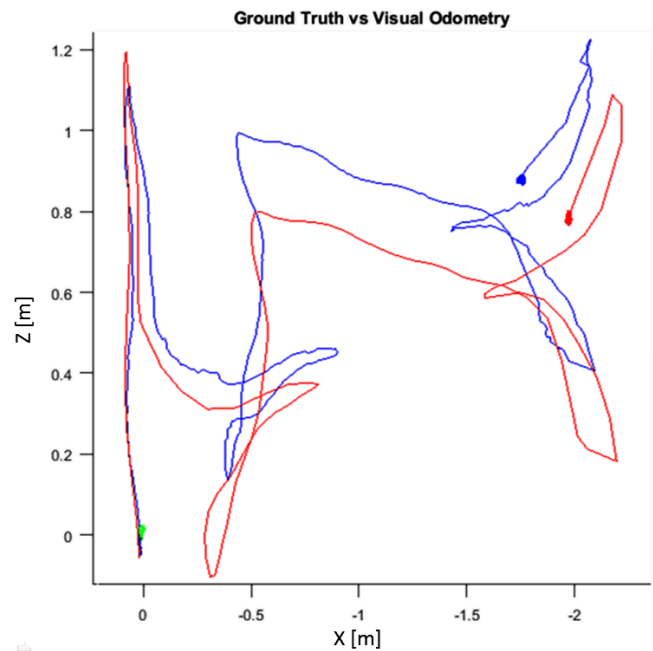

Fig. 10. Trajectory comparison for the image sequency freiburg1/floor. Groundtruth (red) against the computed estimative (blue).

## VI. CONCLUSION

This paper presents an approach to Visual Odometry (VO). The proposed method uses information obtained by an RGB-D camera moving in a static environment and is compared with SfM and ICP, two other approaches to VO frequently used in various applications. Its performance was evaluated in accuracy and processing time on two reference datasets. For both datasets used, the proposed method obtained an accuracy on average 70% better than ICP and 66% better than SfM

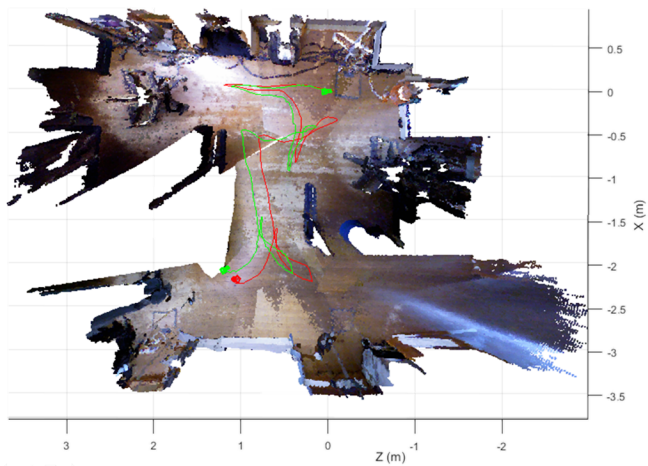

Fig. 11. Top view of the 3D representation of the scene for freiburg1/floor in contrast with the compared trajectory shown on Figure 10. Groundtruth (red) against the computed estimative (green).

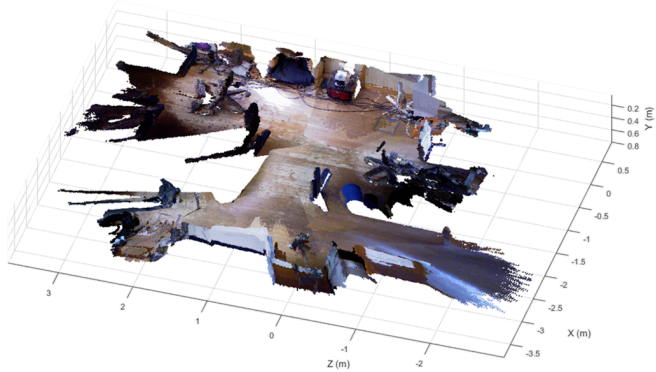

Fig. 12. Perspective view of the 3D representation for freiburg1/floor.

and processing time in average 22% better than ICP and 43% better than SfM.

The proposed algorithm works only with pairs of frames, ignoring all information already processed, which limits its use to static environments and, if used to reconstruct 3D environments, it might not result in the ideal position of the 3D data with respect to the overall reconstruction, generating overlapping information. For a more accurate estimative, one must also add some way to detect loop closure and eventually a global optimization step to obtain a consistent map, thus composing all implementation stages of a Simultaneous Localization and Mapping (SLAM) algorithm.